\documentclass[numbers,preprint,review,12pt]{elsarticle}

\usepackage{amsmath,amsfonts}
\usepackage{amssymb}
\usepackage{amsthm}
\usepackage{mathtools}
\usepackage{algorithmic}
\usepackage{algorithm}
\usepackage{array}
\usepackage{subfig}
\usepackage{textcomp}
\usepackage{stfloats}
\usepackage{url}
\usepackage{verbatim}
\usepackage{graphicx}
\usepackage{multirow}
\usepackage{multicol}
\usepackage{colortbl}
\usepackage[table,xcdraw]{xcolor}
\usepackage{booktabs}
\usepackage{hyperref}
\usepackage{adjustbox}
\usepackage{bm}
\usepackage{makecell}
\allowdisplaybreaks

\journal{Pattern Recognition}

\definecolor{graybg}{rgb}{0.93, 0.93, 0.93}

\theoremstyle{plain}

\theoremstyle{definition}

\theoremstyle{remark}

\begin{document}

\begin{frontmatter}

\title{Robust 3D Alignment of Generative Reconstructions via Partial Monocular Observations}

\author[label1]{\fnref{coequal}Yuchen Zhang}
\ead{13777063701@163.com}

\author[label2]{\fnref{coequal}Luanyuan Dai}
\ead{dailuanyuan@njust.edu.cn}

\author[label3]{\fnref{coequal}Yiwei Wang}
\ead{wang07@seas.upenn.edu}

\author[label4]{Xiwei Xu}
\ead{2024433045@tju.edu.cn}

\author[label5]{Jianing Zhang}
\ead{zhangjianing97313@gmail.com}

\author[label6]{Johnny.r.zhang}
\ead{johnny.r.zhang97@gmail.com}

\author[label7]{Xianhui Meng}
\ead{mengxh@mail.ustc.edu.cn}

\author[label8]{Yanbiao Ma}
\ead{ybma1998@ruc.edu.cn}

\author[label9]{Jiayi Ma}
\ead{jyma2010@gmail.com}

\author[label10]{Xiaoshuai Hao\corref{cor1}}
\ead{haoxiaoshuai714@163.com}

% ===== 脚注部分 =====
\fntext[coequal]{These authors contribute equally to this work.}
\cortext[cor1]{Corresponding author.}

% ===== 机构部分 =====
\affiliation[label1]{
    organization={School of Computer Science, Shanghai Jiao Tong University},
  postcode={200240},
    city={Shanghai},
    country={China}
}

%%%%%%%%%%%%%修改正确%%%%%%%%%%%%%%%%%%%%%%%%%%%%%

\affiliation[label2]{
    organization={School of Computer Science and Engineering, Nanjing University of Science and Technology},
    postcode={210094},
     city={Nanjing},
    country={China}
}

\affiliation[label3]{
    organization={School of Engineering and Applied Science, University of Pennsylvania},
    postcode={19104},
    city={Philadelphia},
    country={USA}
}

\affiliation[label4]{
    organization={School of Tech Electrical and Information Engineering, Tianjin University},
    postcode={300072},
     city={Tianjin},
    country={China}
}

\affiliation[label5]{
    organization={School of Future Information Innovation, Fudan Universiy},
    postcode={200240},
    city={Shanghai},
    country={China}
}

\affiliation[label6]{
    organization={Independent Researcher},
    postcode={33520},
    city={Tampere},
    country={Finland}
}

\affiliation[label7]{
    organization={School of Electronic Engineering and Information Science, University of Science and Technology of China},
    postcode={230026},
    city={Hefei},
    country={China}
}

\affiliation[label8]{
    organization={Gaoling School of Artificial Intelligence, Renmin University of China},
    postcode={100872},
     city={Beijing},
    country={China}
}

\affiliation[label9]{
    organization={School of Robotics, Wuhan University},
    postcode={430072},
     city={Wuhan},
    country={China}
}

\affiliation[label10]{
    organization={Xiaomi EV},
     postcode={100081}, 
     city={Beijing},
    country={China}
}

\begin{abstract}
Aligning generative 3D reconstructions with partial monocular observations is a critical but under-explored challenge in computer vision.
This task is inherently ill-posed due to severe asymmetries between noisy, sparse monocular inputs and dense generative priors, whose scale ambiguity and geometric hallucinations, combined with the lack of initial overlap, render traditional registration pipelines ineffective.
To resolve these issues, we propose a training-free and interpretable geometric alignment framework that grounds generative 3D priors via a 3D similarity transformation (Sim(3)), which can recover accurate metric scale and pose.
Specifically, we introduce an explicit scale factor to resolve metric ambiguity and employ a coarse-to-fine alignment strategy, leveraging geometry-aware descriptors for robust initialization and a decoupled closed-form solver for precision refinement.
In addition, we introduce a \textit{Hallucination Filtering} operation to effectively suppress outliers caused by hallucinated geometry.
To evaluate alignment performance under these extreme conditions, we introduce \textbf{\textit{GenPMOAlign--Where2Place}}, a rigorous benchmark specifically designed for Generative-to-Partial Monocular Observational Alignment.
Experiments demonstrate that our method achieves stable and accurate registration, substantially outperforming both classical geometric pipelines and state-of-the-art learning-based baselines.
Code and the benchmark will be publicly released.

\end{abstract}

% \begin{highlights}
% \item Formalizes canonical-to-metric alignment for robotic perception as a partial-to-complete Sim(3) registration problem.
% \item Develops a robust, training-free framework for accurate metric scale and pose estimation.
% \item Achieves reliable alignment despite generative hallucinations, density mismatch, and little or no overlap.
% \item Establishes the GenPMOAlign-Where2Place benchmark and consistently outperforms existing methods.

% \end{highlights}

\begin{keyword}
3D registration \sep point cloud registration \sep similarity transformation \sep generative reconstruction \sep robotic manipulation
\end{keyword}

\end{frontmatter}

\section{Introduction}
\label{sec:intro}

\begin{figure*}[t]
    \centering
    \includegraphics[width=0.99\textwidth]{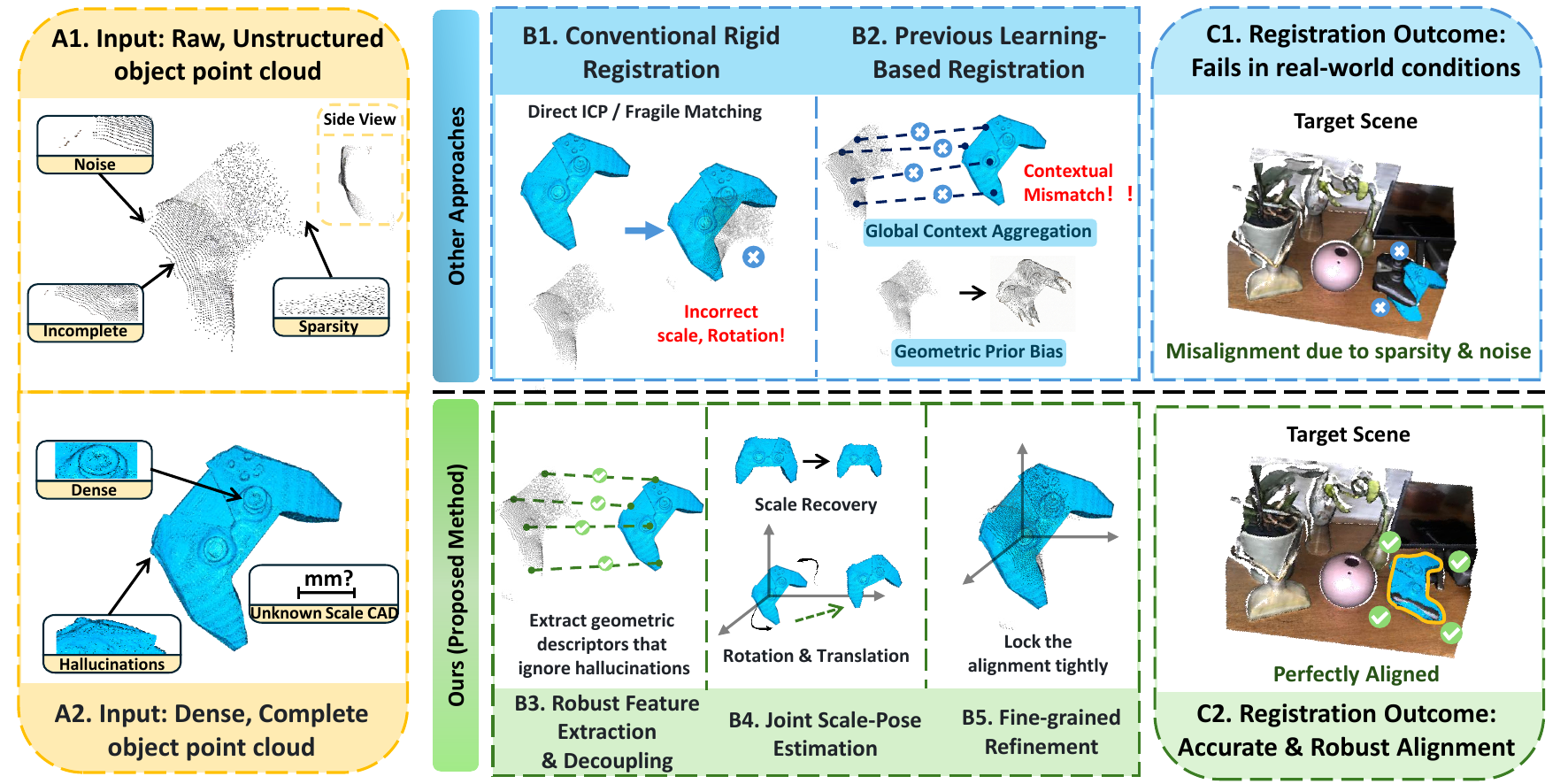}
    \caption{\textbf{Comparison between previous methods and our approach in the generative-to-partial alignment task.}
    The task aims to align noisy, sparse monocular observations (A1) with dense generative priors of unknown scale (A2).
    Traditional rigid registration methods (e.g., ICP) fail due to scale ambiguity and noise (B1), while learning-based methods struggle with contextual mismatch caused by hallucinated geometry (B2).
    Our approach effectively addresses these challenges by incorporating scale recovery, robust feature extraction, and fine-grained refinement, achieving accurate and robust alignment, even in real-world conditions.}
    \label{fig:1}
    \vspace{-3mm}
\end{figure*}

% Registering generative 3D reconstructions in canonical space to metric sensory observations is an emerging yet under-explored problem and a key enabler for geometry-aware robotic manipulation, including grasping \cite{murali2025graspgen,iwase2025zerograsp}, collision avoidance \cite{khatib1986real,wen2023distributed}, and interaction \cite{hogan1984impedance,hao2025mimo}.
% As shown in Fig.~\ref{fig:1}(A1) and (A2), the inputs to this task exhibit pronounced cross-domain discrepancies: the partial point cloud inferred from monocular or depth estimation is typically sparse, noisy, and visibility-limited, whereas the prior produced by the generative model is dense and complete yet often contains geometric hallucinations and lacks a true metric scale; moreover, the two often have little to no initial overlap.
% As depicted in Fig.~\ref{fig:1}(C1), traditional rigid registration methods that rely on overlap and correspondence assumptions are ineffective in this setting.
% Specifically, direct ICP-style approaches \cite{besl1992method,koide2021voxelized,vizzo2023kiss} are brittle under sparse and noisy inputs, often resulting in incorrect scale and rotation, as shown in Fig.~\ref{fig:1}(B1).
% On the other hand, learning-based registration methods, despite leveraging global context, can still fail due to contextual mismatch, prior bias, and density discrepancies, leading to noticeable misalignment in real scenes, as illustrated in Fig.~\ref{fig:1}(B2).
% {\color{red}
Recent advances in embodied intelligence and robotic foundation models have significantly expanded the capability of robots to operate in open-world environments \cite{yuan2025robopoint,hao2025mimo}. To achieve reliable manipulation beyond constrained laboratory settings, robots must acquire a comprehensive understanding of scene geometry and object structure. In grasping, collision avoidance, and physical interaction, they must reason not only about visible surfaces but also about the overall shape, volume, and potential contact regions of objects \cite{fang2020graspnet,murali2025graspgen,khatib1986real}. However, the observations available in real-world scenarios are often limited to monocular or sparse sensory inputs. Occlusions, restricted viewpoints, and sensor noise frequently yield incomplete geometric information, which alone is insufficient for reliable planning and execution.

Unlike traditional reconstruction pipelines that rely on multi-view observations or dedicated depth sensors, recent generative 3D models can recover relatively complete object geometry from a single image or limited observations while plausibly inferring occluded regions \cite{tochilkin2024triposr,chen2025sam,ni2025decompositional,wu20254d}. By providing dense and complete object representations beyond the current sensor field of view, generative priors offer richer geometric constraints for downstream tasks such as grasp planning, motion planning, collision checking, and scene interaction.

Despite their advantages, generative 3D reconstructions cannot be directly deployed in robotic systems. The generated geometry typically resides in a normalized canonical space rather than the metric physical space in which robots perceive and interact with the world \cite{wang2019normalized,chen2024pointreggpt}. Specifically, generative models often produce geometrically plausible shapes without reliable metric scale, and their coordinate systems are internally defined and decoupled from real-world coordinates \cite{tochilkin2024triposr,chen2025sam}. Consequently, the same reconstructed shape may correspond to either a real mug or a miniature toy replica, making it difficult to determine its true physical size and spatial location from the generated model alone. This discrepancy between canonical geometry and metric observations forms a fundamental \emph{canonical-to-metric gap}. For robotic applications, such inconsistency directly affects downstream decision making. Even when the reconstructed shape is visually accurate, scale errors may lead to incorrect gripper configurations, failed collision checking, or inaccurate motion planning. From a geometric perspective, the problem therefore requires not only recovering the coordinate transformation between the canonical and metric spaces, but also estimating the missing absolute scale. Consequently, establishing a consistent mapping between generative priors and metric observations has become a prerequisite for deploying generative 3D models in real-world robotic systems.

Among these discrepancies, metric scale ambiguity is arguably the most fundamental obstacle. Unlike conventional point cloud registration, where both inputs are acquired from physical sensors and naturally share a common metric space \cite{zeng20173dmatch}, generative reconstructions are inherently scale-free. As a result, the alignment problem extends beyond rigid pose estimation and requires jointly recovering scale and pose under severe partial observations \cite{gower1975generalized,umeyama1991least}. Moreover, geometric hallucinations introduced by generative models, together with sparsity, noise, and substantial density mismatch in sensory observations, further increase the difficulty of reliable alignment.

Registering generative 3D reconstructions in canonical space to metric sensory observations is therefore an emerging yet under-explored problem and a key enabler for geometry-aware robotic manipulation, including grasping \cite{murali2025graspgen,iwase2025zerograsp}, collision avoidance \cite{khatib1986real}, and interaction \cite{hao2025mimo}. As shown in Fig.~\ref{fig:1}(A1) and (A2), the inputs to this task exhibit pronounced cross-domain discrepancies: the partial point cloud inferred from monocular or depth estimation \cite{lin2025depth} is typically sparse, noisy, and visibility-limited, whereas the prior produced by the generative model is dense and complete yet often contains geometric hallucinations and lacks a true metric scale \cite{chen2025sam,tochilkin2024triposr}; moreover, the two often have little to no initial overlap. As depicted in Fig.~\ref{fig:1}(C1), traditional rigid registration methods that rely on overlap and correspondence assumptions are ineffective in this setting. Specifically, direct ICP-style approaches \cite{besl1992method,koide2021voxelized,vizzo2023kiss} are brittle under sparse and noisy inputs, often resulting in incorrect scale and rotation, as shown in Fig.~\ref{fig:1}(B1). On the other hand, learning-based registration methods \cite{huang2021predator,qin2023geotransformer}, despite leveraging global context, can still fail due to contextual mismatch, prior bias, and density discrepancies, leading to noticeable misalignment in real scenes, as illustrated in Fig.~\ref{fig:1}(B2).

To address these challenges, we propose a training-free and interpretable geometric alignment framework that grounds generative 3D priors into the metric world via a 3D similarity transformation (Sim(3)), which can recover accurate scale and pose, as presented in Fig.~\ref{fig:1}(B3), (B4), and (B5).
Specifically, we explicitly model the scale factor to resolve metric ambiguity and adopt a coarse-to-fine alignment strategy: we first obtain a robust initialization using geometry descriptors that are insensitive to hallucinations, and then optimize with a decoupled closed-form Sim(3) solver to handle zero-overlap cases and decoupled coordinate systems.
In addition, we introduce a \textit{Hallucination Filtering} operation that restricts optimization to observation-supported correspondences, suppressing outliers induced by hallucinated geometry.
To evaluate alignment under these extreme conditions, we build \textbf{\textit{GenPMOAlign--Where2Place}}, a rigorous benchmark tailored to generative-to-partial monocular observational alignment on cluttered indoor home or office images.
As demonstrated in Fig.~\ref{fig:1}(C2), our approach achieves stable and accurate registration, substantially outperforming classical geometric pipelines and strong learning-based baselines.

Our contributions are threefold:
\begin{enumerate}
    \item We formalize the canonical-to-metric gap for generative 3D models in robotics as a \emph{partial-to-complete} 3D similarity transformation registration problem under unknown scale, density mismatch, and low or zero overlap.
    \item We propose a robust, training-free, and interpretable Sim(3) alignment framework that jointly estimates scale and pose while accounting for generative hallucinations and cross-domain density mismatch, enabling accurate recovery of metric scale and pose.
    \item We construct \textbf{\textit{GenPMOAlign--Where2Place}} to benchmark the cross-domain Sim(3) registration task from generative 3D priors to partial monocular or depth observations, and show that under this extreme regime our method consistently outperforms classical geometric pipelines and learning-based baselines.
\end{enumerate}

\section{Related Work}
\label{sec:related}

\textbf{Classical Point Cloud Registration}
Classical point cloud registration aims to estimate a rigid or similarity transformation to align two or more point clouds, which are generally categorized into local registration and global registration.
Specifically, ICP \cite{besl1992method} alternates correspondence matching and error minimization for fine alignment, but is initialization-sensitive and can get trapped in poor local minima under large pose changes or partial overlap.
To improve geometric modeling, GICP \cite{segal2009generalized} unifies point-to-point and point-to-plane formulations and enhances robustness.
LiTAMIN2 \cite{yokozuka2021litamin2} introduces a symmetric KL-divergence term into the ICP objective to compare local distribution shapes.
Voxelized GICP \cite{koide2021voxelized} and KISS-ICP \cite{vizzo2023kiss} further emphasize efficient implementation and practical deployability.
In recent years, classical frameworks have continued to advance in robust global estimation and efficient search \cite{lin2023robust,huang2024efficient,yan2025turboreg}.
More commonly, correspondence-driven pipelines first establish candidate correspondences using handcrafted descriptors (\textit{e.g.}, FPFH \cite{rusu2009fast}), then obtain an initial pose via robust estimators such as RANSAC \cite{fischler1981random}, and finally refine it with fast global optimization methods like FGR \cite{zhou2016fast}.
Moreover, TEASER \cite{yang2020teaser} and GNC \cite{yang2020graduated} provide stronger robustness and outlier rejection.
In addition, max-clique-based approaches \cite{zhang20233d,fathian2024clipper+,yang2024mac} further improve robustness by selecting mutually consistent correspondence subsets.
% {\color{red}
From a geometric perspective, classical registration methods have achieved remarkable success when both source and target point clouds originate from physical sensors and exhibit similar geometric properties \cite{besl1992method,segal2009generalized}.
Their robustness largely stems from explicit geometric constraints and interpretable optimization objectives.
However, these methods generally assume sufficient overlap, relatively consistent point densities, and reliable initialization \cite{huang2021predator}.
Consequently, their performance often degrades under severe partial observations, cross-domain discrepancies, and density mismatch.

Overall, classical methods are training-free and interpretable, but depend on sufficient overlap, consistent density, and reliable initialization, and thus degrade under cross-domain, partial, or density-mismatched settings.

\textbf{Learning-Based Point Cloud Registration}
Driven by the success of deep learning in 2D vision, point cloud registration shifts from handcrafted features to learned features and correspondences.
Early works learn local descriptors and keypoints (3DMatch \cite{zeng20173dmatch}), while sparse or voxel convolution enables dense feature extraction (FCGF \cite{choy2019fully}, D3Feat \cite{bai2020d3feat}).
PARENet \cite{yao2024pare} boosts rotational robustness with position-aware rotation-equivariant features.
For correspondence modeling, recent methods emphasize global consistency and coarse-to-fine filtering (DGR \cite{choy2020deep}, CoFiNet \cite{yu2021cofinet}, Predator \cite{huang2021predator}), with additional correspondence pruning and confidence estimation (ICC \cite{yuan2024inlier}).
Transformer-based models further strengthen long-range context modeling and improve robustness in low-overlap, cluttered scenes (GeoTransformer \cite{qin2023geotransformer} and follow-ups \cite{chen2024dynamic,chen2024fast}).
% {\color{red}
Compared with classical methods, learning-based approaches seek to learn discriminative geometric representations directly from large-scale data \cite{choy2019fully,huang2021predator}.
By leveraging learned features and contextual reasoning, these methods often demonstrate improved robustness to noise, clutter, and moderate overlap reduction \cite{qin2023geotransformer,yu2021cofinet}.
More recently, foundation-model-based registration methods have emerged as a promising direction for improving cross-domain generalization through large-scale semantic priors \cite{wang2023zeroreg,zheng2025rare,lim2026towards}.

Despite these advances, cross-domain and zero-shot generalization remains challenging.
In particular, SpinNet \cite{ao2021spinnet} and BUFFER \cite{ao2023buffer} target more general descriptors and zero-shot settings.
ZeroReg \cite{wang2023zeroreg} uses 2D foundation-model semantics for object-level zero-shot registration.
RARE \cite{zheng2025rare} leverages foundation models for zero-shot registration and refinement.
BUFFER-X \cite{lim2026towards} is training-free, improving cross-domain generalization with high efficiency.
RAP \cite{pan2025register} further explores generative flow-matching registration that directly produces registered point clouds across diverse scales and sensor setups.

\textbf{Generative 3D Reconstruction for Robotics}
% {\color{red}
Recent advances in generative 3D reconstruction have enabled the recovery of complete object geometry from a single image or sparse observations \cite{tochilkin2024triposr,chen2025sam,ni2025decompositional,wu20254d}.
Leveraging large-scale generative priors, these methods can infer plausible shapes for occluded regions and produce dense, semantically meaningful 3D representations.
Such capabilities have attracted growing attention in robotics, where complete geometric understanding is essential for grasp planning, manipulation, scene interaction, and embodied reasoning \cite{fang2020graspnet,murali2025graspgen,iwase2025zerograsp,yuan2025robopoint}.

Several recent studies have explored the integration of generative 3D models into robotic systems \cite{murali2025graspgen,iwase2025zerograsp}.
By providing shape completion beyond visible observations, generative priors can significantly improve object understanding under severe occlusion and limited viewpoints.
However, unlike CAD models or sensor-acquired point clouds, generative reconstructions are often represented in normalized canonical coordinates and may contain hallucinated structures, scale ambiguity, and topological inconsistencies \cite{tochilkin2024triposr,chen2025sam,ni2025decompositional,wu20254d}.
These characteristics introduce substantial challenges when aligning generative priors with real-world observations and remain largely under-explored in existing registration literature.

\textbf{Scale-Aware Cross-Domain Alignment}
% {\color{red}
Scale-aware alignment has become increasingly important in heterogeneous registration problems, where source and target data are represented in different coordinate systems and possess inconsistent metric scales.
Unlike conventional SE(3) registration, Sim(3)-based alignment jointly estimates scale, rotation, and translation, making it more suitable for cross-domain geometric matching \cite{gower1975generalized,umeyama1991least}.
This capability is particularly important when aligning generative reconstructions with real observations, as generative models generally lack physically meaningful metric information \cite{wang2019normalized,chen2024pointreggpt}.

Heterogeneous data alignment (\textit{e.g.}, CAD-to-depth scans \cite{avetisyan2019scan2cad,lin2024hipose} or generative priors-to-real observations \cite{chen2024pointreggpt}) typically requires solving Sim(3) to jointly estimate rotation, translation, and scale.
Classical solutions build on generalized Procrustes analysis \cite{gower1975generalized} or scale-separation techniques, and methods such as TEASER++ \cite{yang2020teaser} improve robustness under high outlier rates via max-clique constraints, but still rely on accurate correspondences and high geometric fidelity.
Category-level approaches (\textit{e.g.}, NOCS \cite{wang2019normalized}) alleviate scale ambiguity by mapping observations into a normalized canonical space, yet depend on category-specific supervision and dense, high-quality observations, limiting generalization to open-world and cross-category settings.
Hallucinated geometry from generative and foundation models often exhibits scale drift, structural incompleteness, and topological inconsistencies, making uniform Sim(3) solvers prone to failure under severe partiality and density mismatch \cite{ni2025decompositional,wu20254d}.
% {\color{red}
Unlike most prior Sim(3) alignment methods that assume high-fidelity correspondences, dense observations, or learned canonical mappings \cite{avetisyan2019scan2cad,lin2024hipose,wang2019normalized}, the problem considered in this work is substantially more challenging.
The source geometry may contain hallucinated structures, while the target observation is sparse, noisy, and incomplete.
These characteristics jointly introduce scale ambiguity, density mismatch, and low-overlap correspondence uncertainty, motivating the development of a dedicated alignment framework tailored to generative-to-partial observational registration.

Hence, we propose a training-free and interpretable geometric alignment framework that grounds generative 3D priors into the metric world via a Sim(3), thereby recovering accurate scale and pose.

\section{Methodology}
\label{sec:method}

\begin{figure*}[t]
    \centering
    \includegraphics[width=0.99\linewidth]{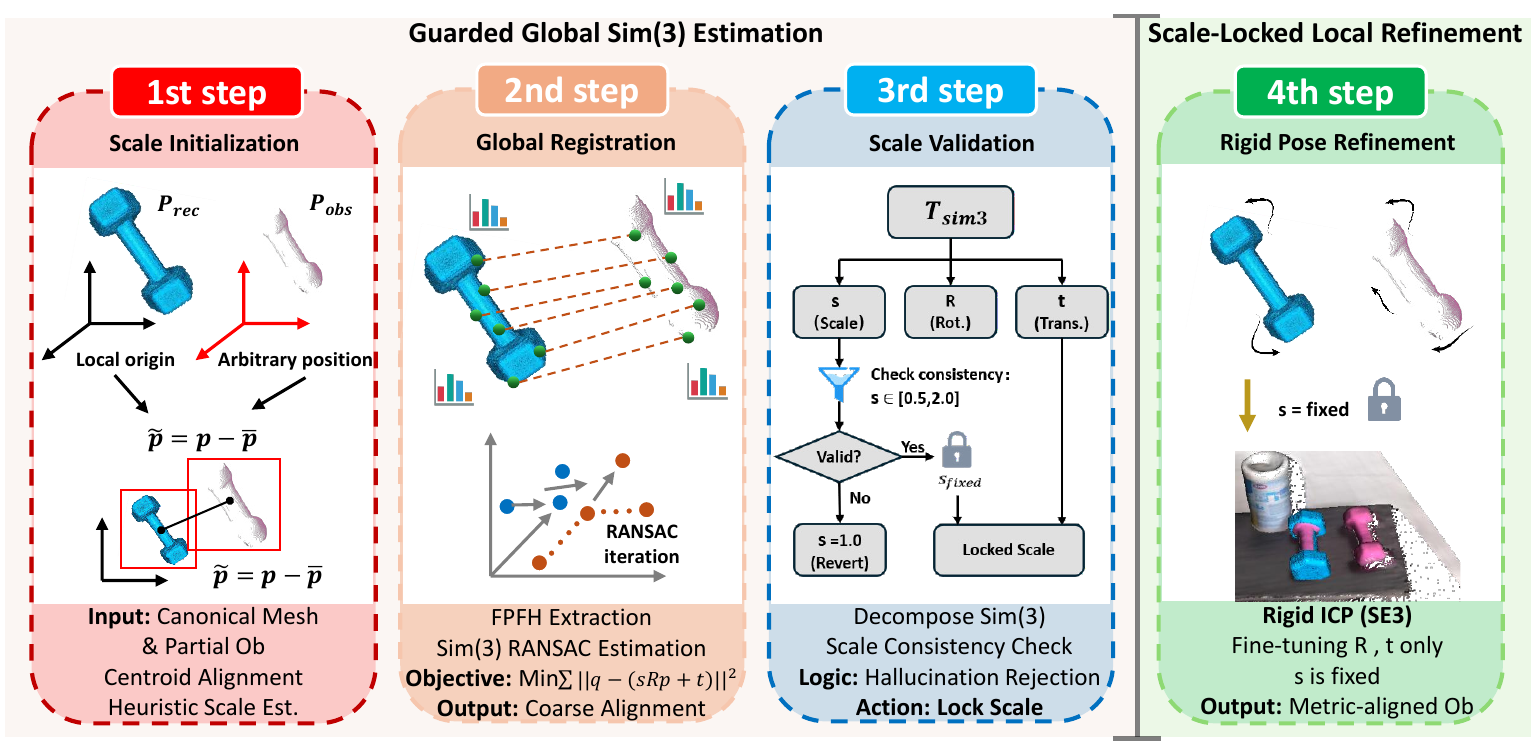}
    \caption{\textbf{Pipeline of the proposed 3D registration method.}
    The process consists of four steps: (1) scale initialization, aligning the origin; (2) global registration, using FPFH and Sim(3) RANSAC for alignment; (3) scale validation, locking the scale if consistent; and (4) rigid pose refinement, applying ICP to refine rotation and translation, producing a metric-aligned object.}
    \label{fig:pipeline}
    \vspace{-4mm}
\end{figure*}

We propose a scale-stabilized Sim(3) registration framework for aligning generative canonical 3D reconstructions with partial monocular observations, as illustrated in Fig.~\ref{fig:pipeline}.
Unlike previous methods that jointly optimize scale and pose until convergence, we decouple scale estimation into a \textit{guarded global initialization} and a \textit{scale-locked local refinement}.
Our approach prevents the generative model from collapsing to fit partial observations and avoids the common shrinking degeneracy.

\subsection{Observation Model and Problem Formulation}
\label{sec:3.1}

The system input consists of a monocular RGB image, a semantic mask, and an estimated depth map $D(u,v)$.
We model the projection from 2D pixels to 3D space using the pinhole camera model with intrinsic matrix $K$.

For valid pixels $(u, v)$ within the target mask, the back-projected point in the camera frame is
\begin{equation}
x_{cam}(u,v) = D(u,v) \cdot K^{-1}
\begin{bmatrix}
u \\ v \\ 1
\end{bmatrix}.
\end{equation}
After back-projection, we transform the point set $X_{cam} = \{x_{cam}^i \in \mathbb{R}^3\}_{i=1}^N$ to obtain the observational point cloud $P_{obs} = \{ q_i \in \mathbb{R}^3 \}_{i=1}^N$ and sample a source point cloud $P_{rec} = \{ p_j \in \mathbb{R}^3 \}_{j=1}^M$ from the canonical mesh generated by SAM 3D \cite{chen2025sam}.
We then estimate a similarity transformation $S = \{s, R, t\} \in \mathrm{Sim}(3)$ that aligns the canonical shape to the metric observation:
\begin{equation}
\min_{s, R, t} \mathcal{J} = \sum_{(i,j) \in \mathcal{C}} \left\| q_i - (s R p_j + t) \right\|^2,
\end{equation}
where $\mathcal{C}$ represents the correspondence set, $s \in \mathbb{R}^+$ is a uniform scale factor, $R \in SO(3)$ is the rotation, and $t \in \mathbb{R}^3$ is the translation.

\subsection{Guarded Global Sim(3) Estimation}
\label{sec:3.2}

We first preprocess the data using FPFH, instead of learning-based descriptors, to keep the method training-free.
However, in this task the significant scale discrepancy between the unit canonical source $P_{rec}$ and the metric observation $P_{obs}$ makes vanilla FPFH matching unstable.
To address this issue, we introduce a heuristic scale factor $s_{init}$ before correspondence search.
We scale $P_{rec}$ by $s_{init}$ so that the support radius for normal estimation and FPFH extraction covers geometrically comparable regions.
The descriptor $\Phi(p)$ then becomes approximately transformation-invariant:
\begin{equation}
s_{init} = \frac{\| \mathrm{extent}(P_{obs}) \|}{\| \mathrm{extent}(P_{rec}) \|},
\end{equation}
\begin{equation}
\Phi(p) \approx \Phi(s_{init} R p + t).
\end{equation}

We further propose a \textit{Hallucination Filtering} operation to suppress unsupported surfaces in the generative prior.
Specifically, we restrict correspondence construction and optimization to regions that are geometrically consistent with the monocular or depth observation, and discard areas with insufficient observation support, preventing hallucinations from biasing scale and pose estimates.

We then use a RANSAC framework to estimate the global alignment from the FPFH correspondences.
For a minimal subset of correspondences, the optimal Sim(3) parameters can be derived in closed form.
The translation is decoupled via centroid alignment, the rotation is solved by SVD, and the scale is obtained by minimizing the residual variance:
\begin{equation}
\Sigma_{pq} = \sum_{(i,j)\in\mathcal{C}} (q_i - \bar q)(p_j - \bar p)^\top,
\end{equation}
\begin{equation}
s^* = \frac{\operatorname{tr}(R^\top \Sigma_{pq})}{\sum_j \| p_j - \bar p \|^2},
\end{equation}
\begin{equation}
R^* \in \arg\max_{R\in SO(3)} \operatorname{tr}(R \Sigma_{pq}),
\end{equation}
\begin{equation}
t^* = \bar q - s^* R^* \bar p.
\end{equation}

In the generative-to-partial setting, unconstrained Sim(3) estimation often yields degenerate scales, for example $s \rightarrow 0$, to minimize error on noisy partial observations.
To avoid this failure mode, we introduce a guarded estimation strategy.
A hypothesis is considered valid only if its estimated scale falls within a physically plausible range relative to the heuristic initialization, namely $s^* \in [\lambda_{\min} s_{init}, \lambda_{\max} s_{init}]$.
If the scale violates this constraint, the hypothesis is rejected during the RANSAC loop, preventing hallucinated geometry from distorting the metric scale estimate.
The accepted solution is denoted by $S_{est} = \{ s^*, R^*, t^* \}$.

\subsection{Scale-Locked Local Refinement}
\label{sec:3.3}

The standard approach typically performs joint optimization of $(s, R, t)$ on the local alignment obtained through RANSAC.
However, we find that allowing the scale $s$ to update during local refinement leads to significant shrinkage of the generated model, because the optimizer attempts to minimize the distance to the incomplete surface of $P_{obs}$.

To address this issue, we propose a \textbf{Scale-Locked Refinement}.
We fix the global scale $\hat{s}$ estimated from the guarded RANSAC stage and restrict the local optimization to the $SE(3)$ manifold.
The refinement objective becomes
\begin{equation}
\min_{R, t} \sum_{(i,j) \in \mathcal{K}} \left\| q_i - (\hat{s} R p_j + t) \right\|^2,
\end{equation}
where $\mathcal{K}$ denotes the local correspondence set used during refinement.

By treating the scale as a hard constraint during the fine-tuning phase, we force rotation and translation to align the observed regions while preserving the global geometric integrity of the hallucinated parts.
The final output is a metric-aligned object representation that is both geometrically stable and robust to hallucinated geometry.

\section{Experiment}
\label{sec:experiments}

\subsection{Benchmarks and Evaluation Metrics}

\textbf{Benchmarks}
We construct \textit{GenPMOAlign--Where2Place}, a benchmark for generative-to-partial monocular alignment.
It consists of 100 cluttered indoor home or office scenes, each with one image and 2--4 paired masks, totaling 338 point cloud pairs.
For Where2Place \cite{yuan2025robopoint}, we use the provided RGB images and object masks.
We reconstruct a complete canonical mesh with SAM 3D \cite{chen2025sam} conditioned on the image and mask, yielding a normalized canonical shape that may contain hallucinated geometry in occluded regions.
We next estimate dense depth from the same image using Depth Anything \cite{lin2025depth} and back-project masked depth to obtain a partial object point cloud along with the surrounding scene point cloud.
Since the generative mesh resides in a normalized canonical space while the observation point cloud is in metric space, this process inherently introduces scale inconsistency.
Furthermore, the presence of partial observations and hallucinated geometry makes the alignment problem ill-posed under rigid transformations, leading to the extreme generative-priors-to-partial-monocular-observations registration setting.

\textbf{Evaluation Metrics}
We evaluate alignment quality using both 2D and 3D metrics.
The 2D metrics include Boundary F-score, Center Drift, and IoU, which measure image-space consistency by comparing the projected silhouette of the aligned reconstruction with the target mask.
Accurate contour alignment is essential for downstream robotic tasks such as grasping and interaction.
The 3D metrics include Chamfer Distance, UniChamfer Distance, Fitness, Normal Consistency, and Depth Mean Absolute Error (Depth MAE), which capture bidirectional geometric error, observation-to-prediction geometric error, overlap quality, normal consistency, and residual inlier error, respectively.
Scale recovery is quantified by Depth MAE, the mean absolute error between the rendered depth of the aligned mesh and the estimated depth map.

\subsection{Generative-to-Partial Monocular Observational Alignment}

\begin{table*}[t]
\centering
\renewcommand{\arraystretch}{1.2} 
\setlength{\tabcolsep}{4pt}
\resizebox{1.0\textwidth}{!}{%
\begin{tabular}{c|l|ccc cccc c} 
\toprule
\multicolumn{2}{c|}{\multirow{2}{*}{\textbf{Method}}} & \multicolumn{3}{c}{\textbf{2D Metrics}} & \multicolumn{5}{c}{\textbf{3D Metrics}} \\ 
\cmidrule(lr){3-5} \cmidrule(lr){6-10} 
\multicolumn{2}{c|}{} & \textbf{Boundary F} $\uparrow$ & \textbf{Center Drift} $\downarrow$ & \textbf{IoU} $\uparrow$ & \textbf{Chamfer} $\downarrow$ & \textbf{Fitness} $\uparrow$ & \textbf{Normal Cons.} $\uparrow$ & \textbf{Depth MAE} $\downarrow$ & \textbf{Unichamfer} $\downarrow$\\
\midrule
\multirow{4}{*}{\rotatebox{90}{\textbf{Traditional}}} 
  & FGR         & 0.2562 & 17.4885 & 0.5030 & 31.5524 & 0.3265 & 0.6638 & 34.7210    & 97.91 \\
  & GICP        & 0.3651 & 14.4621 & 0.5713 & 21.6842 & 0.4004 & 0.7385 & 25.5326    & 22.89 \\
  & ICP P2Plane & 0.3466 & 13.7205 & 0.5725 & 22.7666 & 0.4079 & 0.7298 & 25.9668    & 26.95 \\
  & ICP P2Point & 0.3477 & 10.2559 & 0.6032 & 17.4429 & 0.4420 & 0.7516 & 20.5719    & 18.67 \\
\midrule
\multirow{4}{*}{\rotatebox{90}{\textbf{Deep Learning}}} 
  & Predator    & 0.2599    & 19.3787    & 0.5009    & 28.4041    & \textbf{0.8094}    & 0.7033    & 32.4227    & \textbf{15.5069}    \\
  & GeoTransformer & 0.2155    & 24.9377    & 0.4082    & 40.8592    & 0.6404    & 0.6496    & 41.1876    & 31.4545    \\
  & RAP         & 0.2013 & 24.5546 & 0.3964 & 44.0998 & 0.5284 & 0.6115 & 42.1329    & 38.1860 \\
  & BUFFER-X    & 0.0253 & 113.53 & 0.0683 & 432.61 & 0.0121 & 0.5469 & 361.02    & N/A \\
\midrule
\multicolumn{2}{l|}{\cellcolor{graybg}\textbf{Ours}} &
\cellcolor{graybg}\textbf{0.3926} & \cellcolor{graybg}\textbf{9.1658} & \cellcolor{graybg}\textbf{0.6443} &
\cellcolor{graybg}\textbf{14.8803} & \cellcolor{graybg}0.4777 & \cellcolor{graybg}\textbf{0.7541} & \cellcolor{graybg}\textbf{20.0198} & \cellcolor{graybg}16.02 \\
\bottomrule
\end{tabular}
}
\vspace{1mm}
\caption{\textbf{Quantitative comparison on the GenPMOAlign--Where2Place dataset.} Metrics include Boundary F-score, Center Drift, IoU, Chamfer, Fitness, Normal Consistency, Depth MAE, and UniChamfer. The best result in each column is highlighted in \textbf{bold}.}
\label{tab:registration_results_main}
\end{table*}

\textbf{GenPMOAlign--Where2Place}
Tab.~\ref{tab:registration_results_main} reports quantitative results on GenPMOAlign--Where2Place with the updated protocol, including RAP \cite{pan2025rap} and BUFFER-X \cite{lim2026towards}.
Importantly, all learning-based baselines are evaluated under our unified preprocessing protocol rather than their original off-the-shelf input pipelines: we apply the same mask-guided partial observation construction, pair centering, heuristic scale initialization, and normalized input generation before model-specific inference.
Therefore, the reported differences mainly reflect each method's registration behavior under a common generative-to-partial setting, instead of dataset-specific preprocessing advantages from official implementations.
Our method achieves the best performance among all methods on Boundary F-score, Center Drift, IoU, Chamfer Distance, Normal Consistency, and Depth MAE, while Predator attains higher Fitness and UniChamfer under the same protocol, indicating that high overlap scores alone do not guarantee better silhouette or metric-depth alignment in this setting.
Compared with rigid baselines (FGR/GICP/ICP), the Sim(3)-based formulation better handles canonical-to-metric scale mismatch and yields substantially improved silhouette and geometric alignment quality.

\begin{figure*}[t]
    \centering
    \includegraphics[width=0.95\textwidth]{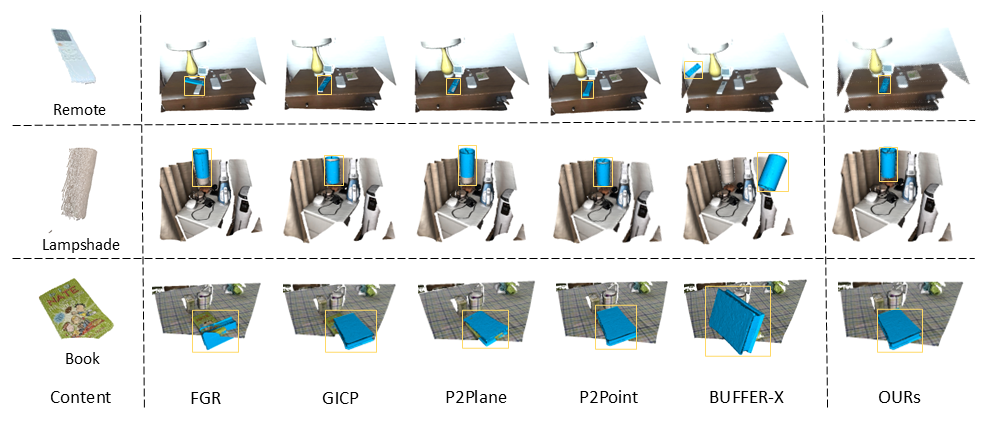}
    \caption{\textbf{Qualitative comparison of representative objects in GenPMOAlign--Where2Place under different registration methods.} The yellow boxes highlight the alignment regions.}
    \label{fig:4}
    \vspace{-4mm}
\end{figure*}

We also observe that several learning-based methods (e.g., Predator and GeoTransformer) can obtain relatively high Fitness, but this does not consistently translate to better global alignment quality in our setting, as reflected by weaker 2D consistency and larger Chamfer/Depth errors.
UniChamfer further reveals this behavior: methods with high overlap scores may still remain less accurate on observation-supported surfaces under severe partial-to-complete mismatch.
Overall, these updated results support our main claim that explicitly modeling scale and constraining registration with observation-consistent geometry provides the most reliable alignment in this challenging regime.
Consistent with the quantitative trends, Fig.~\ref{fig:4} shows that our method produces more stable pose and silhouette alignment on representative instances.

From the 3D metrics, standard rigid registration methods often fail to achieve accurate alignment, yielding higher Chamfer Distance and lower Fitness, primarily due to the scale ambiguity inherent to monocular reconstruction.
Moreover, solvers restricted to $SE(3)$ cannot bridge this scale gap and frequently converge to poor local minima, where object centers align but boundaries drift.
In contrast, our method jointly estimates scale $s$, rotation $R$, and translation $t$ in $Sim(3)$, leading to lower geometric error and more accurate dense alignment.
Beyond scale, partial-to-complete matching induces low or zero overlap: hallucinated regions lack correspondence in single-view observations, making methods such as FGR sensitive to non-overlapping geometry and degrading registration stability.
By enforcing consistency only on observation-supported surfaces, our approach suppresses hallucinations and remains robust under low overlap, which is further reflected by improved 2D metrics, namely higher IoU and Boundary F-score.
Consistent with these quantitative results, Fig.~\ref{fig:4} provides qualitative comparisons on representative object instances.
Traditional methods commonly exhibit pose drift and silhouette misalignment under this cross-domain setting, while our method more reliably recovers plausible scale and pose.

\noindent\textbf{General Registration.}
To test whether our design generalizes beyond the target generative-to-partial setting, we evaluate it on the standard 3DMatch, 3DLoMatch, and KITTI Odometry benchmarks under the established evaluation protocols in prior work.
Following common practice, we report representative baseline results together with our scores.
Our method achieves \textbf{92.0} on 3DMatch and \textbf{36.0} on 3DLoMatch, remaining competitive on generic pairwise registration while being designed for the harder canonical-to-metric alignment problem with partial observations (PARENet reports higher indoor recall on both benchmarks).
On KITTI, our pipeline attains \textbf{99.82\%} registration recall with mean RRE/RTE of \textbf{0.14$^\circ$/0.045\,m} on successful pairs, matching the strongest results reported in BUFFER-X \cite{lim2026towards}.

\begin{table}[!t]
\centering
\scriptsize
\renewcommand{\arraystretch}{1.15}
\setlength{\tabcolsep}{3pt}
\resizebox{\linewidth}{!}{%
\begin{tabular}{lcccccccc}
\toprule
\textbf{Metric} &
\textbf{FPFH+FGR} &
\textbf{FPFH+Quatro} &
\textbf{FPFH+TEASER++} &
\textbf{Predator} &
\textbf{GeoTransformer} &
\textbf{BUFFER} &
\textbf{PARENet} &
\textbf{Ours} \\
\midrule
\textbf{3DMatch} &
62.53 & 8.22 & 52.00 &
90.60 & 92.00 &
92.90 & \textbf{95.00} & 92.00 \\
\textbf{3DLoMatch} &
15.42 & 1.74 & 13.25 &
62.40 & 75.00 &
71.80 & \textbf{80.50} & 36.00 \\
\textbf{KITTI RR} (\%) &
98.74$^\ddagger$ & 99.10$^\ddagger$ & 98.92$^\ddagger$ &
89.9$^\dagger$ & 92.0$^\dagger$ &
99.64$^\ddagger$ & \textbf{99.82$^\ddagger$} & \textbf{99.82} \\
\bottomrule
\end{tabular}
}
\vspace{1mm}
\caption{\textbf{General registration benchmarks.} Registration recall (\%) on 3DMatch, 3DLoMatch, and KITTI under the standard evaluation protocols in prior work; the best result in each row is in \textbf{bold}. $^\dagger$Reported values from the original papers. $^\ddagger$Reported values from BUFFER-X \cite{lim2026towards}.}
\label{tab:general_3dmatch}
\end{table}

\subsection{Real-World Robotic Evaluation}

\begin{table}[t]
  \centering
  \scriptsize
  \setlength{\tabcolsep}{3pt}
  \renewcommand{\arraystretch}{0.95}
  \resizebox{\linewidth}{!}{%
  \begin{tabular}{lcccccccccccc}
    \toprule
    \multirow{1}{*}{Object}
    & \multicolumn{3}{c}{GraspGen}
    & \multicolumn{3}{c}{Ours}
    & \multicolumn{3}{c}{GICP}
    & \multicolumn{3}{c}{ZeroGrasp} \\
    \cmidrule(lr){2-4} \cmidrule(lr){5-7} \cmidrule(lr){8-10} \cmidrule(lr){11-13}
    Category& $S$ & $F$ & $S\%$ & $S$ & $F$ & $S\%$ & $S$ & $F$ & $S\%$ & $S$ & $F$ & $S\%$ \\
    \midrule
    cup   & 6 & 4 & 60 & 10 & 0 & 100 & 7 & 3 & 70 & 8 & 2 & 80 \\
    spoon & 3 & 7 & 30 & 7  & 3 & 70  & 6 & 4 & 60 & 7 & 3 & 70 \\
    plate & 6 & 4 & 60 & 9  & 1 & 90  & 5 & 5 & 50 & 6 & 4 & 60 \\
    hammer& 3 & 7 & 30 & 8  & 2 & 80  & 5 & 5 & 50 & 8 & 2 & 80 \\
    cube  & 5 & 5 & 50 & 10 & 0 & 100 & 7 & 3 & 70 & 9 & 1 & 90 \\
    mouse & 4 & 6 & 40 & 8  & 2 & 70  & 5 & 5 & 50 & 6 & 4 & 60 \\
    \midrule
    Average & - & - & 45.0 & - & - & \textbf{86.7} & - & - & 58.3 & - & - & 73.3 \\
    \bottomrule
  \end{tabular}%
  }
  \vspace{1mm}
  \caption{\textbf{Grasp outcomes predicted by each method on a real robot platform.} We report the numbers of successes (S), failures (F), and the success rate (S\%); best results are in \textbf{bold}.}
  \label{tab:grasp}
  \vspace{-4mm}
\end{table}

\begin{figure*}[t]
    \centering
    \includegraphics[width=0.99\textwidth]{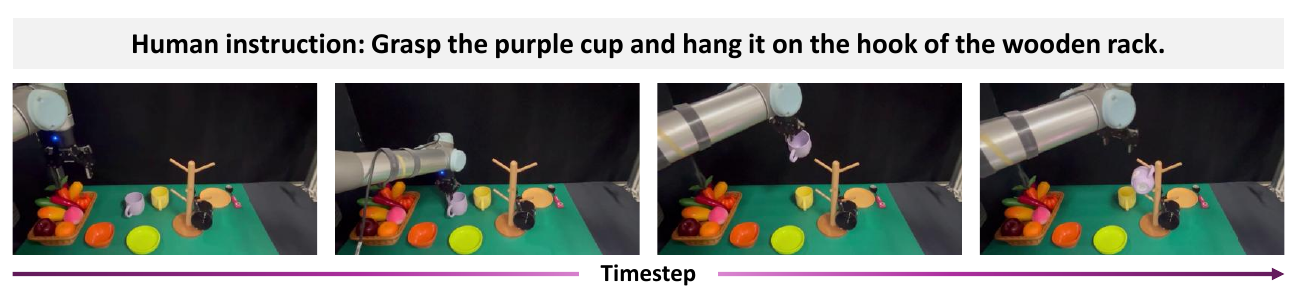}
    \caption{\textbf{Real-world grasping qualitative results.} Given the instruction ``Grasp the purple cup and hang it on the hook of the wooden rack,'' we show a sequence of execution frames over time, illustrating successful grasping and placement on the hook.}
    \label{fig:grasping}
    \vspace{-1mm}
\end{figure*}

We adopt a real-world robot grasping task, which is highly sensitive to metric scale, pose, and local geometry, to perform a task-driven holistic evaluation of generative-to-partial registration and the effectiveness of the proposed method.
Experiments are conducted on a UR5 6-DoF manipulator with a Robotiq 2F-85 gripper and an eye-to-hand monocular RGB camera.
We evaluate six everyday objects (cup, spoon, plate, hammer, Rubik's cube, and mouse), which exhibit thin structures, symmetries, and self-occlusions.
For each object, we reconstruct a sparse point cloud from a single-view observation and execute predicted grasps on the real robot to measure success rates.
GraspGen \cite{murali2025graspgen} predicts grasps directly from the unregistered partial point cloud.
For GICP \cite{segal2009generalized} and our method, we first align the observation to the generative prior and then run GraspGen on the registered point cloud.
ZeroGrasp \cite{iwase2025zerograsp} performs end-to-end shape completion and grasp prediction.
Notably, all baselines except ours and GICP require task-specific training.
Tab.~\ref{tab:grasp} summarizes the grasp success rates of different methods across object categories.
Our method achieves the best average success rate, outperforming both classical geometric registration and learning-based baselines.
It remains reliable on thin and symmetric objects and under noisy, sparse observations, while requiring no task-specific training or fine-tuning.
As shown in Fig.~\ref{fig:grasping}, the system can still produce executable poses and trajectories in a cluttered scene with distractors and challenging contact constraints, achieving stable and precise placement.
Overall, by explicitly aligning generative 3D priors into metric space, our approach bridges generative reconstruction and physical interaction, enabling robust generalization to unseen objects and novel scenes.

\subsection{Ablation Experiments}

\begin{table*}[t]
\centering
\renewcommand{\arraystretch}{1}
\setlength{\aboverulesep}{0pt}
\setlength{\belowrulesep}{0pt}
\resizebox{0.99\textwidth}{!}{%
\begin{tabular}{ll ccc cccc}
\toprule
\textbf{\multirow{2}{*}{ICP}} & \textbf{\multirow{2}{*}{Scale}} & \multicolumn{3}{c}{\textbf{2D Metrics}} & \multicolumn{4}{c}{\textbf{3D Metrics}} \\
\cmidrule(lr){3-5} \cmidrule(lr){6-9}
&& \textbf{Boundary F} $\uparrow$ & \textbf{Center Drift} $\downarrow$ & \textbf{IoU} $\uparrow$ & \textbf{Chamfer} $\downarrow$ & \textbf{Fitness} $\uparrow$ & \textbf{Normal Cons.} $\uparrow$ & \textbf{Depth MAE} $\downarrow$ \\
\midrule
\checkmark &            & 0.3583 & 10.6200 & 0.6040 & 22.7666 & 0.4407 & 0.7453 & 21.8500 \\
           & \checkmark & 0.3428 & 10.6500 & 0.6099 & 17.4429 & 0.3986 & 0.7352 & 25.7100 \\
\rowcolor{graybg}\checkmark & \checkmark & \textbf{0.3926} & \textbf{9.1658} & \textbf{0.6443} & \textbf{14.8803} & \textbf{0.4777} & \textbf{0.7541} & \textbf{20.0198} \\
\bottomrule
\end{tabular}%
}
\vspace{1mm}
\caption{\textbf{Ablation study on GenPMOAlign--Where2Place}. We evaluate ICP refinement and explicit scale estimation.}
\label{tab:2}
\end{table*}

\begin{table*}[t]
\centering
\renewcommand{\arraystretch}{1}
\setlength{\aboverulesep}{0pt}
\setlength{\belowrulesep}{0pt}
\resizebox{0.99\textwidth}{!}{%
\begin{tabular}{l ccc cccc}
\toprule
\textbf{\multirow{2}{*}{Method}} & \multicolumn{3}{c}{\textbf{2D Metrics}} & \multicolumn{4}{c}{\textbf{3D Metrics}} \\
\cmidrule(lr){2-4} \cmidrule(lr){5-8}
& \textbf{Boundary F} $\uparrow$ & \textbf{Center Drift} $\downarrow$ & \textbf{IoU} $\uparrow$ & \textbf{Chamfer} $\downarrow$ & \textbf{Fitness} $\uparrow$ & \textbf{Normal Cons.} $\uparrow$ & \textbf{Depth MAE} $\downarrow$ \\
\midrule
Voxel Size = 50 & \textbf{0.4082} & 9.5558 & 0.6412 & 15.0004 & \textbf{0.4816} & 0.7502 & \textbf{18.4154} \\
Voxel Size = 20 & 0.3629 & 9.9808 & 0.6235 & 15.7202 & 0.4663 & 0.7239 & 21.9356 \\
\rowcolor{graybg} \textbf{Voxel Size = 30 (Ours)} & 0.3926 & \textbf{9.1658} & \textbf{0.6443} & \textbf{14.8803} & 0.4777 & \textbf{0.7541} & 20.0198 \\
\bottomrule
\end{tabular}%
}
\vspace{1mm}
\caption{\textbf{Robustness to voxel downsampling on GenPMOAlign--Where2Place}. We vary the voxel size used for point cloud subsampling (20, 50, and 30 as our default). Results show that our method remains stable across different point-density settings.}
\label{tab:3}
\end{table*}

\begin{table*}[t]
\centering
\renewcommand{\arraystretch}{1}
\setlength{\aboverulesep}{0pt}
\setlength{\belowrulesep}{0pt}
\resizebox{0.99\textwidth}{!}{%
\begin{tabular}{l ccc cccc}
\toprule
\textbf{\multirow{2}{*}{Local Prior}} & \multicolumn{3}{c}{\textbf{2D Metrics}} & \multicolumn{4}{c}{\textbf{3D Metrics}} \\
\cmidrule(lr){2-4} \cmidrule(lr){5-8}
& \textbf{Boundary F} $\uparrow$ & \textbf{Center Drift} $\downarrow$ & \textbf{IoU} $\uparrow$ & \textbf{Chamfer} $\downarrow$ & \textbf{Fitness} $\uparrow$ & \textbf{Normal Cons.} $\uparrow$ & \textbf{Depth MAE} $\downarrow$ \\
\midrule
SAM-3D (default) & 0.3926 & 9.1658 & 0.6443 & \textbf{14.8803} & 0.4777 & 0.7541 & \textbf{20.0198} \\
TripoSR & 0.3518 & \textbf{8.7146} & \textbf{0.6893} & 20.0526 & \textbf{0.9029} & \textbf{0.7689} & 44.3529 \\
\bottomrule
\end{tabular}%
}
\vspace{1mm}
\caption{\textbf{Ablation on local geometry prior on GenPMOAlign--Where2Place.} We replace the canonical mesh source (SAM-3D vs.\ TripoSR) while keeping the same mask-guided partial observation, depth back-projection, and guarded Sim(3) alignment pipeline.}
\label{tab:local_prior_ablation}
\end{table*}

\textbf{Impact of Scale Recovery}
As shown in Tab.~\ref{tab:2}, when we remove explicit scale estimation and only perform ICP refinement in $SE(3)$, the overall alignment quality degrades substantially.
Specifically, 3D geometric error increases and overlap quality drops, while 2D silhouette consistency also deteriorates with larger center drift.
This result indicates that omitting scale recovery leads to global geometric drift and boundary mismatch under scale inconsistency between canonical and metric spaces, confirming that explicit scale estimation is indispensable for bridging the canonical-to-metric gap and achieving stable, accurate alignment.

\textbf{Role of ICP Refinement in Alignment Quality}
As shown in Tab.~\ref{tab:2}, when ICP refinement is removed and only the initial Sim(3) solution is retained, the alignment quality significantly degrades.
Both 3D Chamfer and Fitness worsen, indicating that without ICP refinement, residual errors are not effectively suppressed, leading to a decline in local geometric alignment quality.
Meanwhile, the 2D metrics also decrease, with particularly noticeable drops in silhouette consistency and IoU, demonstrating the importance of ICP refinement in improving local geometric fidelity and overall alignment accuracy.

\textbf{Robustness to Voxel Downsampling}
Tab.~\ref{tab:3} evaluates the impact of voxel downsampling on GenPMOAlign--Where2Place.
As the voxel size varies over a wide range from 20 to 50, our method exhibits only minor fluctuations in both 2D and 3D metrics, indicating strong robustness to point-density changes.
Meanwhile, the default setting with voxel size 30 achieves the best or near-best performance on most key metrics (Tab.~\ref{tab:3}), suggesting a favorable trade-off between preserving geometric detail and suppressing noise.

\textbf{Local Geometry Prior (SAM-3D vs.\ TripoSR).}
Tab.~\ref{tab:local_prior_ablation} studies whether our alignment pipeline is sensitive to the source of canonical geometry.
We swap SAM-3D with TripoSR \cite{tochilkin2024triposr} for single-view mesh reconstruction while keeping all other components unchanged.
TripoSR achieves higher silhouette consistency (IoU of \textbf{0.6893}) and overlap quality (Fitness of \textbf{0.9029}), together with lower center drift (\textbf{8.71\,px}).
SAM-3D remains advantageous in bidirectional Chamfer distance and depth consistency (Depth MAE of \textbf{20.02\,mm} vs.\ \textbf{44.35\,mm}), reflecting that mesh priors with stronger metric-depth alignment are still preferable when dense depth supervision is available.
Overall, the results confirm that our registration core is not tied to a specific reconstruction backend and can ingest alternative single-view priors such as TripoSR, with expected trade-offs between silhouette alignment and metric depth fidelity.

\textbf{Guarded Estimation Ablation (Mechanism Validation).}
We analyze the Sim(3) scale-control mechanism by varying guard range and scale-locking.
\emph{Collapse} reports the fraction of samples with $s^{*}<0.3$ (pathological shrinkage after heuristic scaling); \emph{Clamped} reports the fraction whose RANSAC scale is projected back into the guard interval.
This ablation validates whether guarded estimation suppresses pathological scales, rather than replacing the main benchmark in Tab.~\ref{tab:registration_results_main}.

\begin{table*}[t]
\centering
\renewcommand{\arraystretch}{1}
\setlength{\tabcolsep}{4pt}
\resizebox{0.99\textwidth}{!}{%
\begin{tabular}{l cccc}
\toprule
\textbf{Setting} & \textbf{Guard Range on $s^{*}$} & \textbf{Scale-locked Refine} & \textbf{Collapse} ($s^{*}<0.3$) $\downarrow$ & \textbf{Clamped} $\uparrow$ \\
\midrule
Guarded (default) & $[0.5,1.5]$ & Yes & \textbf{0.0000} & 0.9464 \\
Guarded (tight) & $[0.8,1.2]$ & Yes & \textbf{0.0000} & 0.9732 \\
Guarded (medium) & $[0.5,1.5]$ & Yes & \textbf{0.0000} & 0.9554 \\
Guarded (loose) & $[0.2,2.0]$ & Yes & 0.9643 & 0.9375 \\
Unguarded & $(0,+\infty)$ & Yes & 0.9643 & 0.0000 \\
No-guard + no-lock & $(0,+\infty)$ & No & 0.9911 & 0.0000 \\
Guard + no-lock & $[0.5,1.5]$ & No & \textbf{0.0000} & 0.9821 \\
No-guard + lock & $(0,+\infty)$ & Yes & 0.9732 & 0.0000 \\
\bottomrule
\end{tabular}%
}
\vspace{1mm}
\caption{\textbf{Mechanism-level ablation of guarded scale estimation.} Guarded constraints with moderate/tight ranges effectively prevent scale collapse, while unguarded settings frequently degenerate.}
\label{tab:guarded_ablation}
\end{table*}

\textbf{Key Findings.}
(1) Guarding is critical for preventing pathological scales: collapse drops from 96.43\%--99.11\% (unguarded variants) to 0\% (guarded variants with moderate/tight ranges).
(2) Overly loose bounds behave similarly to unguarded estimation, indicating that an effective guard must be sufficiently tight.
(3) Scale locking alone does not solve collapse; the dominant factor is whether global Sim(3) scale is explicitly constrained.
Overall, this ablation supports the core claim that guarded estimation improves robustness by explicitly filtering implausible scale hypotheses.

\section{Conclusion}

In this study, we propose a novel geometric alignment framework for aligning generative 3D reconstructions with partial monocular observations.
By explicitly modeling scale, using geometry-aware descriptors for robust initialization, and employing refined optimization, our method effectively mitigates scale ambiguity and geometric hallucinations, significantly enhancing registration accuracy.
Through the construction of the \textbf{\textit{GenPMOAlign--Where2Place}} benchmark and experimental validation, the results demonstrate that our approach outperforms traditional geometric methods and state-of-the-art learning-based methods under extreme conditions.
This work provides a stable and reliable solution for applying generative 3D models in robotic perception, offering useful insights and tools for addressing complex 3D reconstruction and alignment problems in computer vision.
Additional visualizations, theoretical guarantees, and detailed derivations are provided in the supplementary material.

\section*{Impact Statement}

This research aims to advance embodied intelligence by developing a unified robotic system capable of both navigation and manipulation tasks.
By demonstrating that cross-task learning can mutually enhance different embodied capabilities, our research contributes to building more efficient and versatile AI agents.
This unified framework approach helps reduce the complexity and cost of deploying robotic systems, making them more applicable to real-world scenarios.
More broadly, this work advances the integration of embodied intelligence capabilities into a single model, paving the way for more powerful and adaptive autonomous systems that better meet human needs in various domains.

\section*{CRediT authorship contribution statement}
Author contribution details will be provided in the final manuscript.
All authors have read and agreed to the published version of the manuscript.

\section*{Code and Data availability}
All code, model weights, and datasets will be made available upon publication.

\section*{Conflicts of Interests}
The authors declare no conflict of interest.

\section*{Generative AI Disclosure}
During the preparation of this manuscript, the authors used ChatGPT (OpenAI) solely for language editing and clarity improvement of the text.
All content generated with the assistance of this tool was reviewed and edited by the authors, who take full responsibility for the final manuscript.
No generative AI tools were used to create, modify, or enhance any images in this work.
All figures are original and were generated by the authors in accordance with Elsevier's policy on the use of generative AI for images.

\bibliographystyle{elsarticle-num}
\bibliography{main}

@article{lin2025depth,
  title={Depth anything 3: Recovering the visual space from any views},
  author={Lin, Haotong and Chen, Sili and Liew, Junhao and Chen, Donny Y and Li, Zhenyu and Shi, Guang and Feng, Jiashi and Kang, Bingyi},
  journal={arXiv preprint arXiv:2511.10647},
  year={2025}
}

@inproceedings{besl1992method,
  title={Method for registration of 3-D shapes},
  author={Besl, Paul J and McKay, Neil D},
  booktitle={Sensor fusion IV: control paradigms and data structures},
  volume={1611},
  pages={586--606},
  year={1992},
  organization={Spie}
}

@article{pan2025register,
  title={Register Any Point: Scaling 3D Point Cloud Registration by Flow Matching},
  author={Pan, Yue and Sun, Tao and Zhu, Liyuan and Nunes, Lucas and Armeni, Iro and Behley, Jens and Stachniss, Cyrill},
  journal={arXiv preprint arXiv:2512.01850},
  year={2025}
}

@inproceedings{segal2009generalized,
  title={Generalized-icp},
  author={Segal, Aleksandr and Haehnel, Dirk and Thrun, Sebastian},
  booktitle={Robotics: science and systems},
  pages={435},
  year={2009},
  organization={Seattle, WA}
}

@inproceedings{koide2021voxelized,
  title={Voxelized GICP for fast and accurate 3D point cloud registration},
  author={Koide, Kenji and Yokozuka, Masashi and Oishi, Shuji and Banno, Atsuhiko},
  booktitle={2021 IEEE international conference on robotics and automation (ICRA)},
  pages={11054--11059},
  year={2021},
  organization={IEEE}
}

@article{vizzo2023kiss,
  title={Kiss-icp: In defense of point-to-point icp--simple, accurate, and robust registration if done the right way},
  author={Vizzo, Ignacio and Guadagnino, Tiziano and Mersch, Benedikt and Wiesmann, Louis and Behley, Jens and Stachniss, Cyrill},
  journal={IEEE Robotics and Automation Letters},
  volume={8},
  number={2},
  pages={1029--1036},
  year={2023},
  publisher={IEEE}
}

@inproceedings{rusu2009fast,
  title={Fast point feature histograms (FPFH) for 3D registration},
  author={Rusu, Radu Bogdan and Blodow, Nico and Beetz, Michael},
  booktitle={2009 IEEE international conference on robotics and automation},
  pages={3212--3217},
  year={2009},
  organization={IEEE}
}

@article{fischler1981random,
  title={Random sample consensus: a paradigm for model fitting with applications to image analysis and automated cartography},
  author={Fischler, Martin A and Bolles, Robert C},
  journal={Communications of the ACM},
  volume={24},
  number={6},
  pages={381--395},
  year={1981},
  publisher={ACM New York, NY, USA}
}

@inproceedings{zhou2016fast,
  title={Fast global registration},
  author={Zhou, Qian-Yi and Park, Jaesik and Koltun, Vladlen},
  booktitle={European conference on computer vision},
  pages={766--782},
  year={2016},
  organization={Springer}
}

@article{yang2020teaser,
  title={Teaser: Fast and certifiable point cloud registration},
  author={Yang, Heng and Shi, Jingnan and Carlone, Luca},
  journal={IEEE Transactions on Robotics},
  volume={37},
  number={2},
  pages={314--333},
  year={2020},
  publisher={IEEE}
}

@article{yang2020graduated,
  title={Graduated non-convexity for robust spatial perception: From non-minimal solvers to global outlier rejection},
  author={Yang, Heng and Antonante, Pasquale and Tzoumas, Vasileios and Carlone, Luca},
  journal={IEEE Robotics and Automation Letters},
  volume={5},
  number={2},
  pages={1127--1134},
  year={2020},
  publisher={IEEE}
}

@inproceedings{zhang20233d,
  title={3D registration with maximal cliques},
  author={Zhang, Xiyu and Yang, Jiaqi and Zhang, Shikun and Zhang, Yanning},
  booktitle={Proceedings of the IEEE/CVF conference on computer vision and pattern recognition},
  pages={17745--17754},
  year={2023}
}

@article{fathian2024clipper+,
  title={Clipper+: a fast maximal clique algorithm for robust global registration},
  author={Fathian, Kaveh and Summers, Tyler},
  journal={IEEE Robotics and Automation Letters},
  volume={9},
  number={4},
  pages={3562--3569},
  year={2024},
  publisher={IEEE}
}

@article{yang2024mac,
  title={Mac: Maximal cliques for 3d registration},
  author={Yang, Jiaqi and Zhang, Xiyu and Wang, Peng and Guo, Yulan and Sun, Kun and Wu, Qiao and Zhang, Shikun and Zhang, Yanning},
  journal={IEEE Transactions on Pattern Analysis and Machine Intelligence},
  year={2024},
  publisher={IEEE}
}

@inproceedings{zeng20173dmatch,
  title={3dmatch: Learning the matching of local 3d geometry in range scans},
  author={Zeng, Andy and Song, Shuran and Nie{\ss}ner, Matthias and Fisher, Matthew and Xiao, Jianxiong and Funkhouser, T},
  booktitle={CVPR},
  pages={4},
  year={2017}
}

@inproceedings{choy2019fully,
  title={Fully convolutional geometric features},
  author={Choy, Christopher and Park, Jaesik and Koltun, Vladlen},
  booktitle={Proceedings of the IEEE/CVF international conference on computer vision},
  pages={8958--8966},
  year={2019}
}

@inproceedings{bai2020d3feat,
  title={D3feat: Joint learning of dense detection and description of 3d local features},
  author={Bai, Xuyang and Luo, Zixin and Zhou, Lei and Fu, Hongbo and Quan, Long and Tai, Chiew-Lan},
  booktitle={Proceedings of the IEEE/CVF conference on computer vision and pattern recognition},
  pages={6359--6367},
  year={2020}
}

@inproceedings{choy2020deep,
  title={Deep global registration},
  author={Choy, Christopher and Dong, Wei and Koltun, Vladlen},
  booktitle={Proceedings of the IEEE/CVF conference on computer vision and pattern recognition},
  pages={2514--2523},
  year={2020}
}

@inproceedings{yu2021cofinet,
  title={Cofinet: Reliable coarse-to-fine correspondences for robust pointcloud registration},
  author={Yu, Hao and Li, Fu and Saleh, Mahdi and Busam, Benjamin and Ilic, Slobodan},
  booktitle={Advances in Neural Information Processing Systems},
  volume={34},
  pages={23872--23884},
  year={2021}
}

@inproceedings{huang2021predator,
  title={Predator: Registration of 3d point clouds with low overlap},
  author={Huang, Shengyu and Gojcic, Zan and Usvyatsov, Mikhail and Wieser, Andreas and Schindler, Konrad},
  booktitle={Proceedings of the IEEE/CVF Conference on computer vision and pattern recognition},
  pages={4267--4276},
  year={2021}
}

@inproceedings{yuan2024inlier,
  title={Inlier confidence calibration for point cloud registration},
  author={Yuan, Yongzhe and Wu, Yue and Fan, Xiaolong and Gong, Maoguo and Miao, Qiguang and Ma, Wenping},
  booktitle={Proceedings of the IEEE/CVF Conference on Computer Vision and Pattern Recognition},
  pages={5312--5321},
  year={2024}
}

@article{qin2023geotransformer,
  title={Geotransformer: Fast and robust point cloud registration with geometric transformer},
  author={Qin, Zheng and Yu, Hao and Wang, Changjian and Guo, Yulan and Peng, Yuxing and Ilic, Slobodan and Hu, Dewen and Xu, Kai},
  journal={IEEE Transactions on Pattern Analysis and Machine Intelligence},
  volume={45},
  number={8},
  pages={9806--9821},
  year={2023},
  publisher={IEEE}
}

@inproceedings{chen2024dynamic,
  title={Dynamic cues-assisted transformer for robust point cloud registration},
  author={Chen, Hong and Yan, Pei and Xiang, Sihe and Tan, Yihua},
  booktitle={Proceedings of the IEEE/CVF Conference on Computer Vision and Pattern Recognition},
  pages={21698--21707},
  year={2024}
}

@inproceedings{chen2024fast,
  title={Fast and robust point cloud registration with tree-based transformer},
  author={Chen, Guangyan and Wang, Meiling and Yang, Yi and Yuan, Li and Yue, Yufeng},
  booktitle={2024 IEEE International Conference on Robotics and Automation (ICRA)},
  pages={773--780},
  year={2024},
  organization={IEEE}
}

@inproceedings{ao2021spinnet,
  title={Spinnet: Learning a general surface descriptor for 3d point cloud registration},
  author={Ao, Sheng and Hu, Qingyong and Yang, Bo and Markham, Andrew and Guo, Yulan},
  booktitle={Proceedings of the IEEE/CVF conference on computer vision and pattern recognition},
  pages={11753--11762},
  year={2021}
}

@inproceedings{ao2023buffer,
  title={Buffer: Balancing accuracy, efficiency, and generalizability in point cloud registration},
  author={Ao, Sheng and Hu, Qingyong and Wang, Hanyun and Xu, Kai and Guo, Yulan},
  booktitle={Proceedings of the IEEE/CVF conference on computer vision and pattern recognition},
  pages={1255--1264},
  year={2023}
}

@inproceedings{zheng2025rare,
  title={RARE: Refine Any Registration of Pairwise Point Clouds via Zero-Shot Learning},
  author={Zheng, Chengyu and Huang, Jin and Chen, Honghua and Wei, Mingqiang},
  booktitle={Proceedings of the IEEE/CVF International Conference on Computer Vision},
  pages={26549--26558},
  year={2025}
}

@article{gower1975generalized,
  title={Generalized procrustes analysis},
  author={Gower, John C},
  journal={Psychometrika},
  volume={40},
  number={1},
  pages={33--51},
  year={1975},
  publisher={Springer-Verlag}
}

@inproceedings{wang2019normalized,
  title={Normalized object coordinate space for category-level 6d object pose and size estimation},
  author={Wang, He and Sridhar, Srinath and Huang, Jingwei and Valentin, Julien and Song, Shuran and Guibas, Leonidas J},
  booktitle={Proceedings of the IEEE/CVF conference on computer vision and pattern recognition},
  pages={2642--2651},
  year={2019}
}

@inproceedings{ni2025decompositional,
  title={Decompositional neural scene reconstruction with generative diffusion prior},
  author={Ni, Junfeng and Liu, Yu and Lu, Ruijie and Zhou, Zirui and Zhu, Song-Chun and Chen, Yixin and Huang, Siyuan},
  booktitle={Proceedings of the Computer Vision and Pattern Recognition Conference},
  pages={6022--6033},
  year={2025}
}

@inproceedings{wu20254d,
  title={4D-Fly: Fast 4D Reconstruction from a Single Monocular Video},
  author={Wu, Diankun and Liu, Fangfu and Hung, Yi-Hsin and Qian, Yue and Zhan, Xiaohang and Duan, Yueqi},
  booktitle={Proceedings of the Computer Vision and Pattern Recognition Conference},
  pages={16663--16673},
  year={2025}
}

@article{khatib1986real,
  title={Real-time obstacle avoidance for manipulators and mobile robots},
  author={Khatib, Oussama},
  journal={The international journal of robotics research},
  volume={5},
  number={1},
  pages={90--98},
  year={1986},
  publisher={Sage Publications Sage CA: Thousand Oaks, CA}
}

@inproceedings{yuan2025robopoint,
  title={RoboPoint: A Vision-Language Model for Spatial Affordance Prediction in Robotics},
  author={Yuan, Wentao and Duan, Jiafei and Blukis, Valts and Pumacay, Wilbert and Krishna, Ranjay and Murali, Adithyavairavan and Mousavian, Arsalan and Fox, Dieter},
  booktitle={Conference on Robot Learning},
  pages={4005--4020},
  year={2025},
  organization={PMLR}
}

@inproceedings{fang2020graspnet,
  title={Graspnet-1billion: A large-scale benchmark for general object grasping},
  author={Fang, Hao-Shu and Wang, Chenxi and Gou, Minghao and Lu, Cewu},
  booktitle={Proceedings of the IEEE/CVF conference on computer vision and pattern recognition},
  pages={11444--11453},
  year={2020}
}

@inproceedings{yao2024pare,
  title={PARE-Net: Position-aware rotation-equivariant networks for robust point cloud registration},
  author={Yao, Runzhao and Du, Shaoyi and Cui, Wenting and Tang, Canhui and Yang, Chengwu},
  booktitle={European Conference on Computer Vision},
  pages={287--303},
  year={2024},
  organization={Springer}
}

@article{lim2026towards,
  title={Towards Zero-Shot Point Cloud Registration Across Diverse Scales, Scenes, and Sensor Setups},
  author={Lim, Hyungtae and Seo, Minkyun and Carlone, Luca and Park, Jaesik},
  journal={arXiv preprint arXiv:2601.02759},
  year={2026}
}

@article{wang2023zeroreg,
  title={Zeroreg: Zero-shot point cloud registration with foundation models},
  author={Wang, Weijie and Ren, Wenqi and Mei, Guofeng and Ren, Bin and Huang, Xiaoshui and Poiesi, Fabio and Sebe, Nicu and Lepri, Bruno},
  journal={arXiv preprint arXiv:2312.03032},
  year={2023}
}

@article{murali2025graspgen,
  title={Graspgen: A diffusion-based framework for 6-dof grasping with on-generator training},
  author={Murali, Adithyavairavan and Sundaralingam, Balakumar and Chao, Yu-Wei and Yuan, Wentao and Yamada, Jun and Carlson, Mark and Ramos, Fabio and Birchfield, Stan and Fox, Dieter and Eppner, Clemens},
  journal={arXiv preprint arXiv:2507.13097},
  year={2025}
}

@inproceedings{iwase2025zerograsp,
  title={ZeroGrasp: Zero-Shot Shape Reconstruction Enabled Robotic Grasping},
  author={Iwase, Shun and Irshad, Muhammad Zubair and Liu, Katherine and Guizilini, Vitor and Lee, Robert and Ikeda, Takuya and Amma, Ayako and Nishiwaki, Koichi and Kitani, Kris and Ambrus, Rares and others},
  booktitle={Proceedings of the Computer Vision and Pattern Recognition Conference},
  pages={17405--17415},
  year={2025}
}

@article{chen2025sam,
  title={Sam 3d: 3dfy anything in images},
  author={Chen, Xingyu and Chu, Fu-Jen and Gleize, Pierre and Liang, Kevin J and Sax, Alexander and Tang, Hao and Wang, Weiyao and Guo, Michelle and Hardin, Thibaut and Li, Xiang and others},
  journal={arXiv preprint arXiv:2511.16624},
  year={2025}
}

@inproceedings{yan2025turboreg,
  title={Turboreg: Turboclique for robust and efficient point cloud registration},
  author={Yan, Shaocheng and Shi, Pengcheng and Zhao, Zhenjun and Wang, Kaixin and Cao, Kuang and Wu, Ji and Li, Jiayuan},
  booktitle={Proceedings of the IEEE/CVF International Conference on Computer Vision},
  pages={26371--26381},
  year={2025}
}

@article{huang2024efficient,
  title={Efficient and robust point cloud registration via heuristics-guided parameter search},
  author={Huang, Tianyu and Li, Haoang and Peng, Liangzu and Liu, Yinlong and Liu, Yun-Hui},
  journal={IEEE Transactions on Pattern Analysis and Machine Intelligence},
  volume={46},
  number={10},
  pages={6966--6984},
  year={2024},
  publisher={IEEE}
}

@inproceedings{lin2023robust,
  title={Robust point cloud registration with geometry-based transformation invariant descriptor},
  author={Lin, Jianjie and Rickert, Markus and Wen, Long and Hu, Yingbai and Knoll, Alois},
  booktitle={2023 IEEE/RSJ International Conference on Intelligent Robots and Systems (IROS)},
  pages={7163--7170},
  year={2023},
  organization={IEEE}
}

@article{hao2025mimo,
  title={MiMo-Embodied: X-Embodied Foundation Model Technical Report},
  author={Hao, Xiaoshuai and Zhou, Lei and Huang, Zhijian and Hou, Zhiwen and Tang, Yingbo and Zhang, Lingfeng and Li, Guang and Lu, Zheng and Ren, Shuhuai and Meng, Xianhui and others},
  journal={arXiv preprint arXiv:2511.16518},
  year={2025}
}

@article{umeyama1991least,
  title={Least-squares estimation of transformation parameters between two point patterns},
  author={Umeyama, Shinji},
  journal={IEEE Transactions on pattern analysis and machine intelligence},
  volume={13},
  number={4},
  pages={376--380},
  year={1991},
  publisher={IEEE}
}

@inproceedings{avetisyan2019scan2cad,
  title={Scan2cad: Learning cad model alignment in rgb-d scans},
  author={Avetisyan, Armen and Dahnert, Manuel and Dai, Angela and Savva, Manolis and Chang, Angel X and Nie{\ss}ner, Matthias},
  booktitle={Proceedings of the IEEE/CVF Conference on computer vision and pattern recognition},
  pages={2614--2623},
  year={2019}
}

@inproceedings{lin2024hipose,
  title={Hipose: Hierarchical binary surface encoding and correspondence pruning for rgb-d 6dof object pose estimation},
  author={Lin, Yongliang and Su, Yongzhi and Nathan, Praveen and Inuganti, Sandeep and Di, Yan and Sundermeyer, Martin and Manhardt, Fabian and Stricker, Didier and Rambach, Jason and Zhang, Yu},
  booktitle={Proceedings of the IEEE/CVF Conference on Computer Vision and Pattern Recognition},
  pages={10148--10158},
  year={2024}
}

@inproceedings{chen2024pointreggpt,
  title={Pointreggpt: Boosting 3d point cloud registration using generative point-cloud pairs for training},
  author={Chen, Suyi and Xu, Hao and Li, Haipeng and Luo, Kunming and Liu, Guanghui and Fu, Chi-Wing and Tan, Ping and Liu, Shuaicheng},
  booktitle={European Conference on Computer Vision},
  pages={272--289},
  year={2024},
  organization={Springer}
}

@inproceedings{yokozuka2021litamin2,
  title={LiTAMIN2: Ultra light LiDAR-based SLAM using geometric approximation applied with KL-divergence},
  author={Yokozuka, Masashi and Koide, Kenji and Oishi, Shuji and Banno, Atsuhiko},
  booktitle={2021 IEEE international conference on robotics and automation (ICRA)},
  pages={11619--11625},
  year={2021},
  organization={IEEE}
}

@article{tochilkin2024triposr,
  title={{TripoSR}: Fast 3D Object Reconstruction from a Single Image},
  author={Tochilkin, Dmitry and Pankratz, David and Liu, Zexiang and Huang, Zixuan and Letts, Adam and Li, Yangguang and Liang, Ding and Laforte, Christian and Jampani, Varun and Cao, Yan-Pei},
  journal={arXiv preprint arXiv:2403.02151},
  year={2024}
}

\clearpage

\end{document}